\documentclass[letterpaper, 10 pt, conference]{ieeeconf}  %

\IEEEoverridecommandlockouts                              
\usepackage[utf8]{inputenc}
\usepackage[T1]{fontenc}
\usepackage{graphicx}
\usepackage{booktabs}
\usepackage{multirow}
\usepackage{xcolor}
\usepackage{array}
\usepackage{capt-of}
\usepackage{adjustbox}
\usepackage{tabularx}

\usepackage{amssymb}
\usepackage{placeins}
\usepackage{makecell}
\usepackage{amsmath}

\usepackage[table]{xcolor}
\definecolor{rblue}{rgb}{0,0.5,1}
\definecolor{awesome}{rgb}{1.0, 0.13, 0.32}
\definecolor{hollywoodcerise}{rgb}{0.96, 0.0, 0.63}
\definecolor{lasallegreen}{rgb}{0.03, 0.47, 0.19}
\definecolor{hanpurple}{rgb}{0.32, 0.09, 0.98}
\definecolor{green(pigment)}{rgb}{0.0, 0.65, 0.31}

\makeatletter
\let\NAT@parse\undefined
\makeatother
\usepackage[pagebackref=false, breaklinks=true, colorlinks, bookmarks=false]{
        hyperref
}
\hypersetup{
        colorlinks=true,
        linkcolor={red},
        citecolor={hanpurple},
        urlcolor={magenta}
}

\newcommand{\platformicon}[1]{%
    \raisebox{-0.22\height}{%
        \includegraphics[height=1.00em]{#1}%
    }%
}

\newcommand{\platformrowicon}[1]{%
    \makebox[1.48em][c]{%
        \raisebox{-0.16\height}{%
            \includegraphics[
                width=1.28em,
                height=0.92em,
                keepaspectratio
            ]{#1}%
        }%
    }%
}

\newcommand{\platformname}[2]{%
    \platformrowicon{#1}\hspace{0.12em}#2%
}

\title{\LARGE \bf
EmbPASS: Towards Cross-Embodiment Open Panoramic Segmentation
}
\author{Pujun Guo$^{1}$, Yuanfan Zheng$^{1}$, Fei Teng$^{1}$, Mengfei Duan$^{1}$, Guoqiang Zhao$^{1}$,\\Yuheng Zhang$^{1}$, Kai Luo$^{1}$, and Kailun Yang$^{1,2,\dag}$%
\thanks{This work was supported in part by the National Natural Science Foundation of China (Grant No. 62473139 and No. 62388101), in part by the Hunan Provincial Research and Development Project (Grant No. 2025QK3019), and in part by the State Key Laboratory of Autonomous Intelligent Unmanned Systems (the opening project number ZZKF2025-2-10).}%
\thanks{$^{1}$The authors are with the School of Artificial Intelligence and Robotics, Hunan University, China (email: kailun.yang@hnu.edu.cn).}%
\thanks{$^{2}$The author is also with the National Engineering Research Center of Robot Visual Perception and Control Technology, Hunan University, China.}
\thanks{$^{\dag}$Corresponding author: Kailun Yang.}
}

\let\oldtwocolumn\twocolumn
\renewcommand\twocolumn[1][]{%
    \oldtwocolumn[{#1}{
        \begin{center}
            \vskip -2ex

            \includegraphics[
                width=0.99\textwidth
            ]{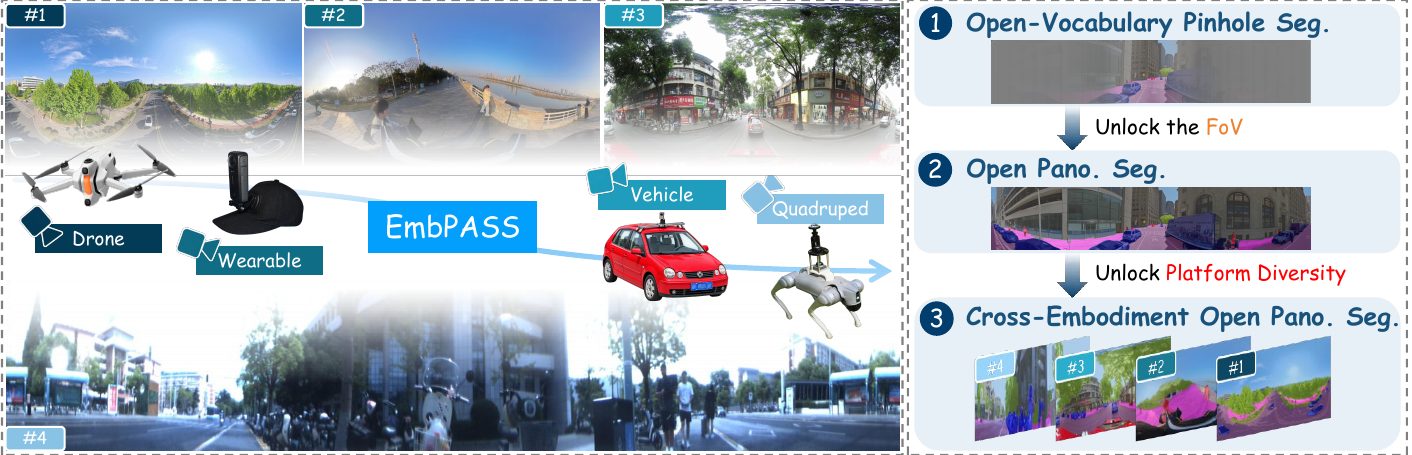}

            \vspace{-0.6ex}
            \makebox[0.99\textwidth][c]{%
                \makebox[0.64\textwidth][c]{%
                    {\footnotesize (a)}%
                }%
                \makebox[0.35\textwidth][c]{%
                    {\footnotesize (b)}%
                }%
            }
    
            \vskip -1.8ex

            \captionof{figure}{
                \hspace*{-0.5em}%
                \textbf{Overview of EmbPASS and cross-embodiment open panoramic segmentation.}
                (a) EmbPASS is a multi-platform panoramic semantic segmentation benchmark spanning four representative embodied platforms: \emph{Vehicle}, \emph{Drone}, \emph{Wearable}, and \emph{Quadruped}, capturing diverse embodied viewpoints.
                (b) Open panoramic segmentation extends open-vocabulary perception from FoV-limited pinhole images to 360{\textdegree} panoramic observations, while cross-embodiment open panoramic segmentation further extends it across heterogeneous embodied platforms.
            }
            \label{fig:teaser}
        
            \vskip 1ex
        \end{center}
    }]
}

\begin{document}

\maketitle
\thispagestyle{empty}
\pagestyle{empty}

\begin{abstract}
Panoramic images provide a complete \(360^\circ\) field of view, enabling comprehensive scene understanding for embodied perception. However, heterogeneous embodied platforms exhibit substantial differences in observation viewpoints and spatial layouts, giving rise to cross-embodiment observation shifts that pose additional challenges to consistent and reliable panoramic perception, while systematic studies of this problem remain limited. To bridge this gap, we introduce a new task, termed \textbf{Cross-Embodiment Open Panoramic Segmentation}. Meanwhile, we establish \textbf{EmbPASS}, a multi-platform panoramic semantic segmentation benchmark spanning \emph{Vehicle}, \emph{Drone}, \emph{Wearable}, and \emph{Quadruped} platforms under a unified semantic taxonomy, providing a testbed for systematically studying cross-embodiment panoramic perception. We further propose \textbf{EPONet}, an open-vocabulary panoramic semantic segmentation network that integrates \textbf{Relation-Aware Metric Adapter (RAMA)} and \textbf{Content-Adaptive Semantic Transfer (CAST)} to enhance spatial modeling and semantic transfer under heterogeneous embodied observations. Extensive experiments show that EPONet achieves the best platform-balanced performance on EmbPASS with \(35.82\%\) mIoU, outperforming the strongest baseline by \(1.10\%\), while remaining competitive on existing panoramic segmentation benchmarks. The source code and EmbPASS benchmark will be made publicly available at \url{https://github.com/guopj1/EmbPASS}.
\end{abstract}

\section{Introduction}
Environmental perception is fundamental to autonomous robots operating in complex and unconstrained real-world environments, where reliable scene understanding supports navigation, planning, and decision-making~\cite{vlmaps,conceptgraphs}. In recent years, panoramic vision has attracted increasing attention in robotic scene understanding due to its complete \(360^\circ\) Field of View (FoV) and more comprehensive environmental observations~\cite{gao2022review,ai2025survey}. Despite the inherent projection distortions of panoramic imagery, panoramic scene understanding has continued to advance and has progressively expanded toward more open perception settings~\cite{trans4pass,dppass,datr,acd4ops}. 

However, systematic studies of panoramic semantic segmentation across heterogeneous embodied platforms remain relatively limited~\cite{trans4pass,dppass,acd4ops,ops}.
In practical robotic deployments, differences in embodied platform configurations and sensor mounting conditions often result in distinct observation heights and viewpoints~\cite{eventfly,perspective3d}, leading to substantial differences in panoramic appearance and spatial layout. Such differences give rise to cross-embodiment observation shifts, posing additional challenges to consistent and reliable panoramic perception. Meanwhile, acquiring dense panoramic annotations for each embodied platform is costly and becomes increasingly impractical as the diversity of embodied platforms continues to grow.

To enable panoramic scene understanding across diverse embodied platforms and address the above challenges, we introduce a new task, termed \textbf{Cross-Embodiment Open Panoramic Segmentation}. Accordingly, the model is trained only on pinhole images from a label-rich source domain and directly performs open-vocabulary semantic segmentation on panoramic images captured by unseen embodied platforms, without requiring target-platform annotations. As illustrated in Fig.~\ref{fig:teaser}(b), OPS~\cite{ops} extends the FoV to \(360^\circ\) panoramic observations, while our task further considers heterogeneous embodied platforms to unlock platform diversity.

To systematically study this task, we construct \textbf{EmbPASS}, a multi-platform panoramic semantic segmentation benchmark for heterogeneous embodied platforms, as shown in Fig.~\ref{fig:teaser}(a). EmbPASS covers multiple representative platforms under a unified semantic taxonomy, providing a common testbed for evaluating cross-embodiment open panoramic segmentation. Compared with existing panoramic benchmarks~\cite{stanford2d3d,matterport3d,wildpass,densepass}, EmbPASS explicitly incorporates platform diversity, enabling systematic analysis of platform-dependent observation variations. EmbPASS further reveals clear platform-dependent differences in the spherical spatial organization of scene semantics. Since ERP projection distortion is inherently latitude-dependent, such semantic layout variations introduce additional geometric challenges for cross-embodiment panoramic perception.

Accordingly, we propose \textbf{EPONet}, an open-vocabulary panoramic semantic segmentation network for heterogeneous embodied platforms. To accommodate the diverse local distortion conditions encountered in heterogeneous embodied observations, we introduce Relation-Aware Metric Adapter (RAMA), which predicts metric parameters from content representations and directional neighborhood relations and uses them to constrain the sampling support, enabling structured spatial aggregation under diverse local geometric distortions. Meanwhile, to better exploit transferable pretrained semantics under varying observation conditions, we propose Content-Adaptive Semantic Transfer (CAST), which adaptively reorganizes and transfers intermediate CLIP~\cite{clip} representations according to the current observation.

Extensive experiments on the established EmbPASS benchmark and existing panoramic segmentation benchmarks verify the effectiveness of EPONet. EPONet achieves the best platform-balanced performance on EmbPASS, reaching $35.82\%$ mIoU and outperforming the strongest competing method by $1.10\%$. It also remains competitive on the public Stanford2D3D~\cite{stanford2d3d} and DensePASS~\cite{densepass} benchmarks.

In summary, our main contributions are as follows:
\begin{itemize}
    \item We introduce \textbf{Cross-Embodiment Open Panoramic Segmentation} and further establish \textbf{EmbPASS}, a multi-platform panoramic semantic segmentation benchmark for systematically studying this task.
    \item We propose \textbf{EPONet}, a cross-embodiment open panoramic segmentation network that integrates RAMA and CAST for effective panoramic perception across heterogeneous embodied platforms.
    \item Extensive experiments on EmbPASS and existing panoramic benchmarks demonstrate the effectiveness of EPONet, which achieves the best platform-balanced performance on EmbPASS while remaining competitive on public panoramic benchmarks.
\end{itemize}

\section{Related Work}

\subsection{Open-Vocabulary Semantic Segmentation}
Open-Vocabulary Semantic Segmentation (OVSS) leverages vision-language models such as CLIP~\cite{clip} to segment an open set of text-defined categories, but their image-level pre-training does not directly provide the spatial precision required for dense prediction. Existing training-based methods adapt vision-language models to dense prediction either by constructing and aggregating pixel-level image-text similarities~\cite{catseg,sed,s2corr} or by classifying category-agnostic mask proposals through region-text matching~\cite{openseg,ovseg,fcclip,maskadapter}. 
Within the mask-based paradigm, SAN~\cite{san} attaches a lightweight side network to a frozen CLIP backbone and decouples mask proposal generation from CLIP-based region recognition.
Despite these advances, most OVSS methods are designed for and evaluated primarily on perspective images.

\subsection{Panoramic Segmentation and Cross-Platform Perception}
Panoramic semantic segmentation provides dense scene understanding over a complete \(360^\circ\) field of view~\cite{gao2022review,ai2025survey}. Existing studies mitigate projection distortions through spherical geometry modeling~\cite{sphericalcnn,sgat4pass,relsf4pass} and distortion-aware spatial modeling~\cite{distortionconv}, while reducing reliance on panoramic annotations through unsupervised domain adaptation~\cite{trans4pass,dppass,datr,densepass,goodsam,omnisam} and its multi-source and source-free extensions~\cite{msda4pass,dapass}.
Recent studies have also extended panoramic segmentation beyond conventional closed-set prediction to broader task settings~\cite{panoos,edapseg}. In particular, OPS and subsequent work extend open-vocabulary segmentation to panoramic imagery, with models trained on FoV-restricted pinhole source images and directly evaluated on unseen panoramic domains~\cite{acd4ops,ops}.

Meanwhile, cross-view and cross-platform perception have attracted increasing attention in robotic vision.
Existing studies transfer perceptual knowledge across substantially different viewpoints and platforms through cross-view adaptation and domain alignment~\cite{eventfly,crovia}; other works improve generalization across platforms and sensing configurations through geometric alignment and feature-level alignment, platform-aware modeling, and explicit modeling of camera configuration priors~\cite{perspective3d,coin3d,embodiedgrounding}. 
Despite advances in both panoramic segmentation and cross-platform perception, open-vocabulary panoramic segmentation across heterogeneous embodied platforms remains largely unexplored.

\begin{table}[t]
    \centering
    \caption{
        Comparison of panoramic semantic segmentation benchmarks.
        Abbreviations:
        Stat. (stationary),
        Sys. (system),
        Platf. (platform),
        \protect\platformicon{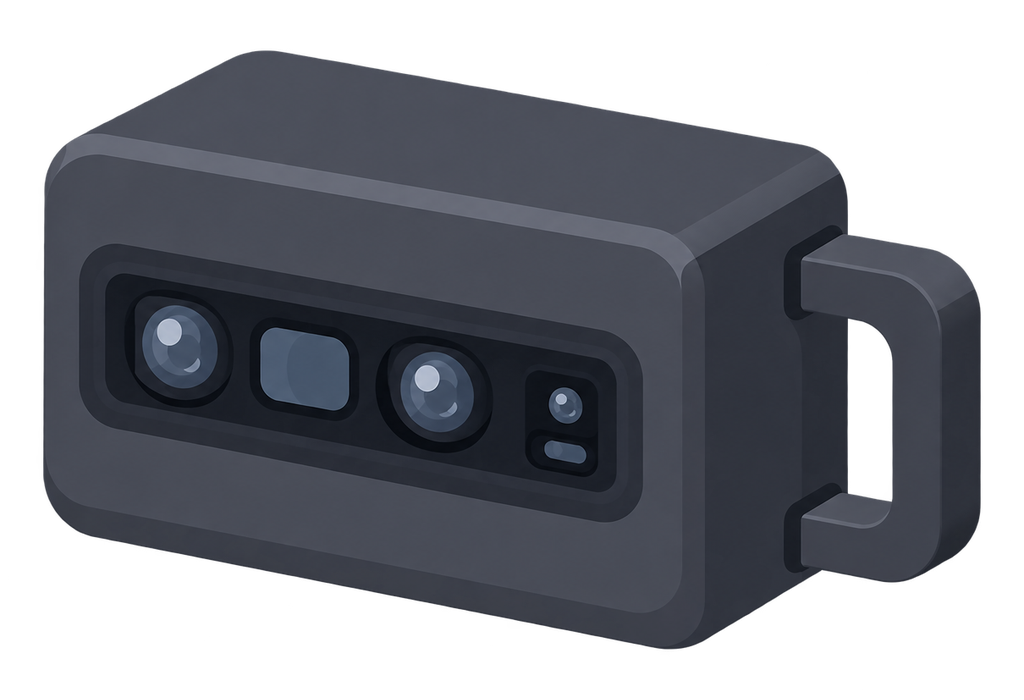} (stationary scanner),
        \protect\platformicon{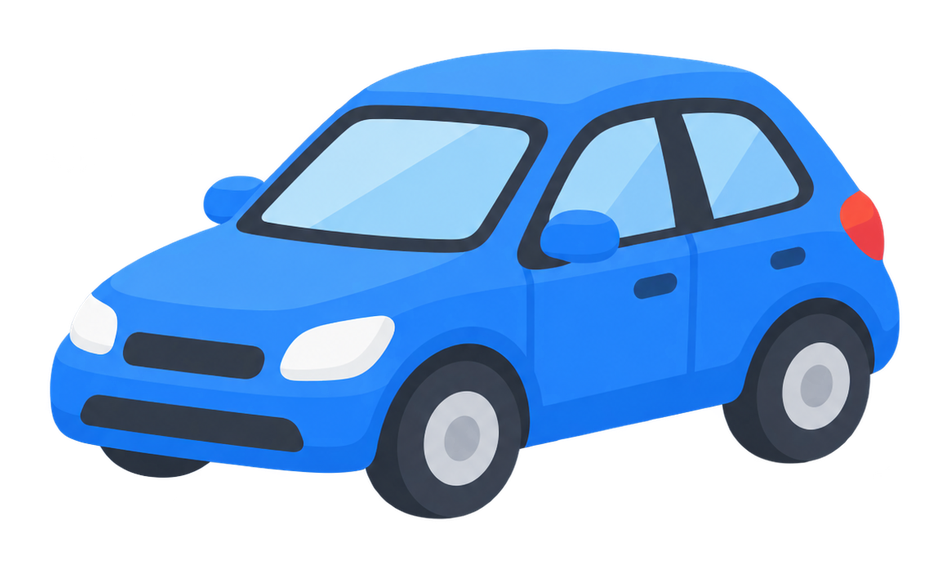} (vehicle),
        \protect\platformicon{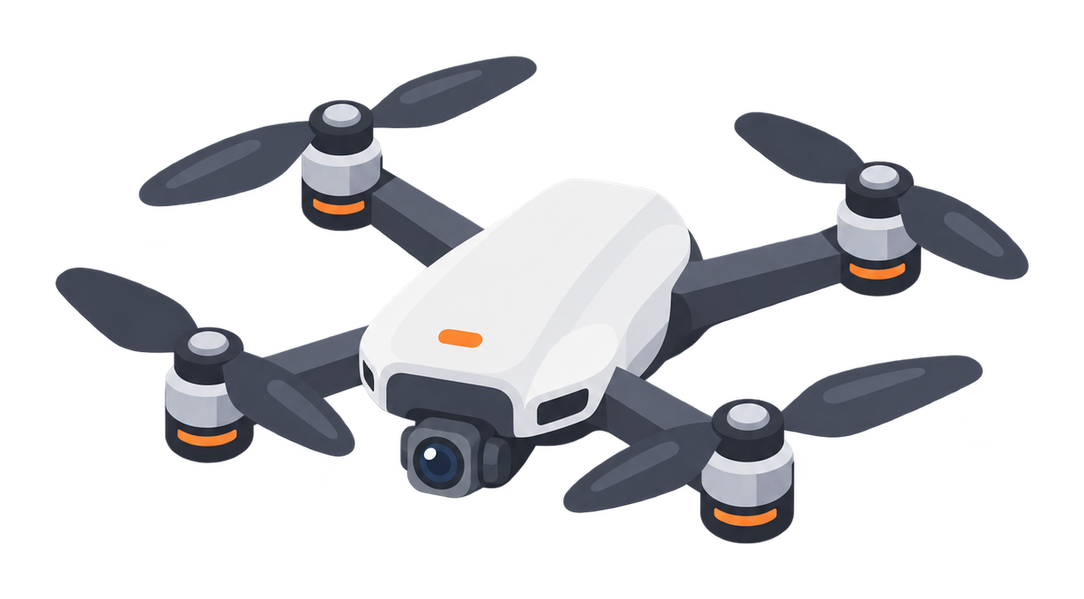} (drone),
        \protect\platformicon{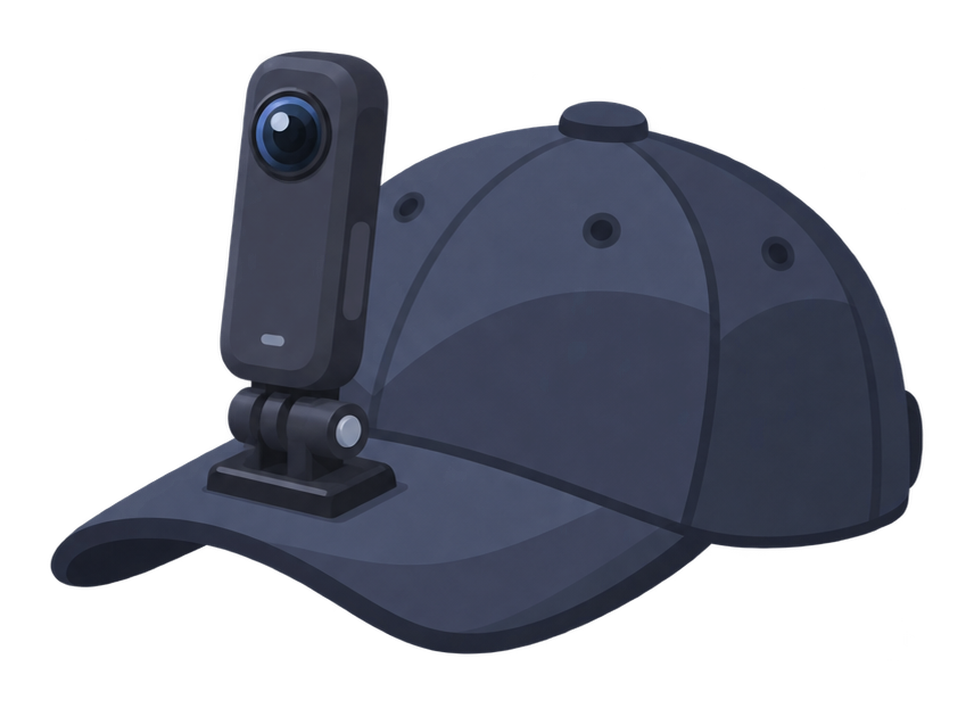} (wearable), 
        \protect\platformicon{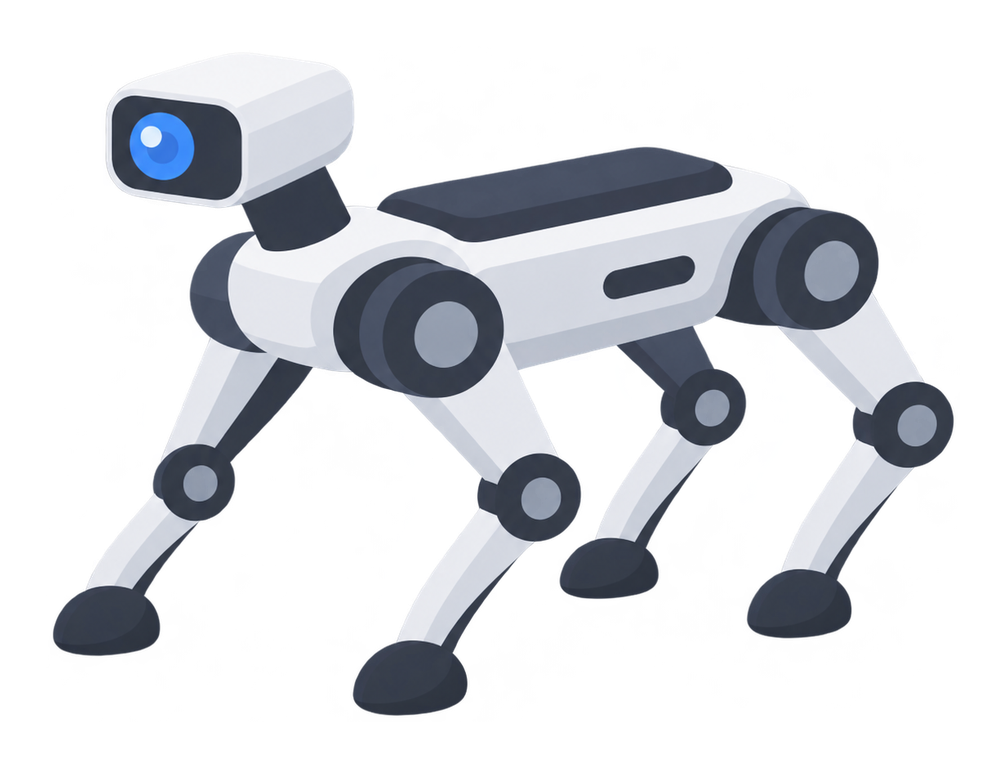} (quadruped).
    }
    \label{tab:benchmark_comparison}
    \vskip-1ex

    \footnotesize
    \setlength{\tabcolsep}{2.2pt}
    \renewcommand{\arraystretch}{1.06}

    \resizebox{\columnwidth}{!}{%
    \begin{tabular}{@{}lccccc@{}}
        \toprule
        Dataset
        & Scene
        & \# Eval.
        & \# Cls.
        & Acq. Platform
        & Platform Coverage \\
        \midrule

        Stanford2D3D~\cite{stanford2d3d}
        & Indoor
        & 1,413
        & 13
        & Stat. scanner
        & \platformicon{figs/scanner_icon.png} \\

        Matterport3D~\cite{matterport3d}
        & Indoor
        & 772
        & 20
        & Stat. scanner
        & \platformicon{figs/scanner_icon.png} \\

        DensePASS~\cite{densepass}
        & Outdoor
        & 100
        & 19
        & Street-view sys.
        & \platformicon{figs/car_icon.png} \\

        WildPASS~\cite{wildpass}
        & Outdoor
        & 500
        & 8
        & Street-view sys.
        & \platformicon{figs/car_icon.png} \\

        \midrule

        \textbf{EmbPASS}
        & Outdoor
        & 1,000
        & 19
        & \textbf{Embodied platf.}
        &
        \platformicon{figs/car_icon.png}\hspace{0.12em}
        \platformicon{figs/drone_icon.png}\hspace{0.12em}
        \platformicon{figs/wearable_icon.png}\hspace{0.12em}
        \platformicon{figs/quad_icon.png}
        \\
        \bottomrule
    \end{tabular}%
    }
    \vskip-4ex
\end{table}

\begin{figure*}[t]
    \centering

    \begin{tabular}{
        @{}c@{\hspace{0.005\textwidth}}c@{}
    }

    \begin{minipage}[t]{0.50\textwidth}
        \vspace{0pt}
        \centering

        {\footnotesize
        \setlength{\tabcolsep}{1.2pt}
        \renewcommand{\arraystretch}{0.94}

        \setlength{\aboverulesep}{0.20ex}
        \setlength{\belowrulesep}{0.20ex}

        \begin{tabular*}{\linewidth}{
            @{\extracolsep{\fill}}lccc@{}
        }
            \toprule
            Platform
            & \# Images
            & Resolution
            & Latitude Range \\
            \midrule

            \platformname{figs/car_icon.png}{Vehicle}
            & 250
            & $2560 \times 972$
            & $[-47^\circ,\,90^\circ]$ \\

            \platformname{figs/drone_icon.png}{Drone}
            & 250
            & $2560 \times 1280$
            & $[-90^\circ,\,90^\circ]$ \\

            \platformname{figs/wearable_icon.png}{Wearable}
            & 250
            & $2560 \times 1280$
            & $[-90^\circ,\,90^\circ]$ \\

            \platformname{figs/quad_icon.png}{Quadruped}
            & 250
            & $2560 \times 320$
            & $[-10^\circ,\,35^\circ]$ \\

            \bottomrule
        \end{tabular*}
        }

        \vspace{0.25pt}

        \makebox[\linewidth][c]{%
            {\footnotesize
            \textbf{(a)} Platform-level statistics}%
        }

        \vspace{-1pt}

        \includegraphics[
            width=\linewidth
        ]{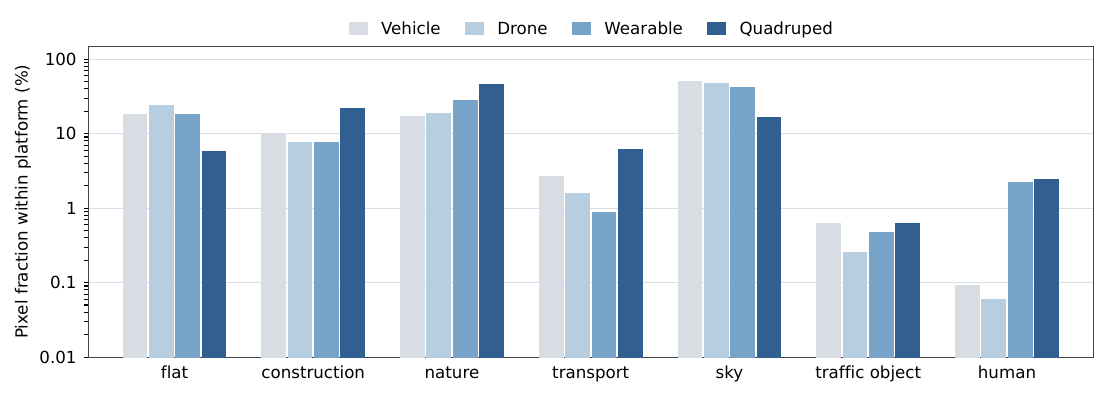}

        \vspace{-0.25pt}

        \makebox[\linewidth][c]{%
            {\footnotesize
            \textbf{(b)} Platform-specific semantic composition}%
        }

    \end{minipage}

    &

    \begin{minipage}[t]{0.48\textwidth}
        \vspace{0pt}
        \centering

        \includegraphics[
            width=\linewidth
        ]{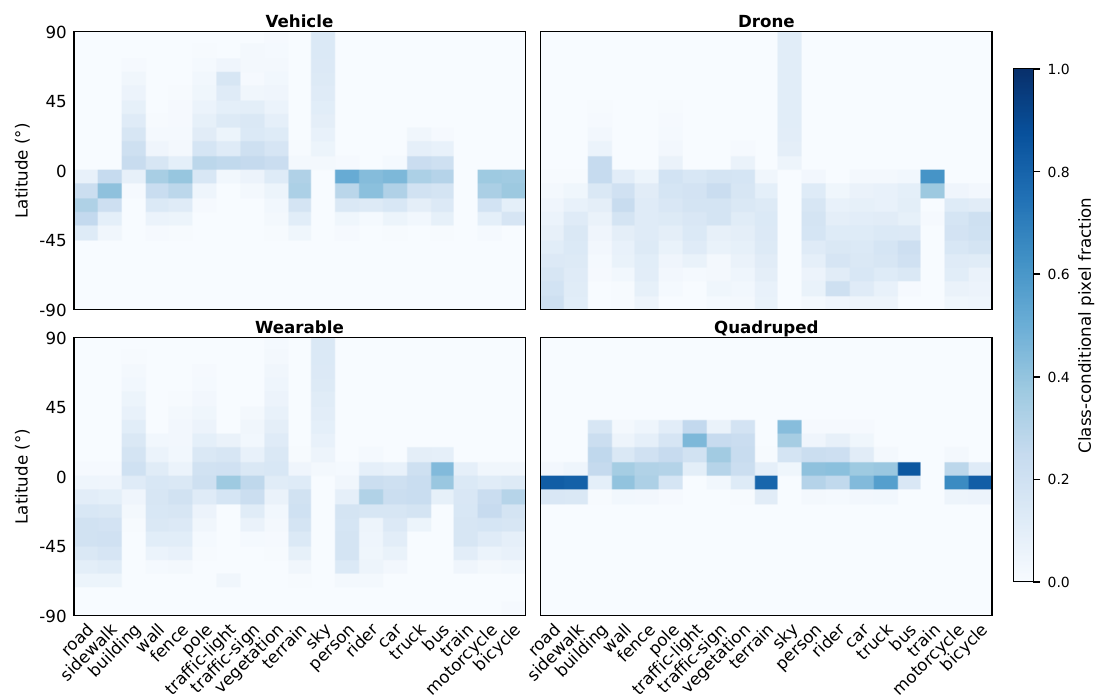}

        {\footnotesize
        \textbf{(c)} Class-wise spherical-latitude distributions
        \par}

    \end{minipage}

    \end{tabular}

    \caption{
        \textbf{Dataset statistics and cross-platform analysis of EmbPASS.}
        (a) Platform-level statistics, including the number of images,
        spatial resolution, and spherical latitude coverage.
        (b) Platform-specific semantic composition, measured by the
        fraction of valid labeled pixels belonging to each semantic group.
        (c) Class-wise spherical latitude distributions across the four
        platform subsets, where color intensity denotes the normalized
        pixel fraction within each spherical latitude bin.
    }
    \label{fig:benchmark_characteristics}
    \vskip-3ex
\end{figure*}

\subsection{Adaptive Spatial Sampling}
Deformable convolutions adapt local sampling locations by learning spatially
varying offsets, thereby improving their ability to model complex geometric
variations~\cite{dcn}. Subsequent variants introduce modulation mechanisms~\cite{dcnv2}, grouped spatial aggregation~\cite{dcnv3}, and more efficient implementations~\cite{dcnv4}. In panoramic segmentation, distortion-aware spatial sampling has been explored through geometry-guided and deformable mechanisms to accommodate spatially non-uniform projection distortions~\cite{trans4pass,ops,sgat4pass,distortionconv}.
In contrast to deformable operators that directly predict pointwise sampling offsets, recent work parameterizes adaptive sampling through low-dimensional, signal-dependent local metrics, thereby constructing structured and content-adaptive sampling supports~\cite{metricconv}.

\section{Established Benchmark}
\subsection{Benchmark Overview}
Existing panoramic segmentation benchmarks cover diverse scene types and domains, but provide limited coverage of heterogeneous embodied platforms. To bridge this gap, we establish EmbPASS, a multi-platform panoramic semantic segmentation benchmark spanning \emph{Vehicle}, \emph{Drone}, \emph{Wearable}, and \emph{Quadruped} under a unified semantic taxonomy, as summarized in Table~\ref{tab:benchmark_comparison}. 
EmbPASS is designed for cross-embodiment open panoramic segmentation and provides a testbed for evaluating model generalization to panoramic observations from heterogeneous embodied platforms.

\subsection{Dataset Construction}
We consider four representative embodied platforms: \emph{Vehicle}, \emph{Drone}, \emph{Wearable}, and \emph{Quadruped}.
Specifically, the \emph{Vehicle} subset is curated from vehicle-mounted panoramic imagery in Spheriverse~\cite{teng2026spheriverse}, while the \emph{Wearable} and \emph{Quadruped} subsets are curated from panoramic imagery captured by a head-mounted camera in WanderDream~\cite{liu2026wanderdream} and a camera mounted on a quadruped robot in PanoMMOcc~\cite{zhao2026panommocc}, respectively. 
For the \emph{Drone} subset, we collect outdoor panoramic sequences using an Antigravity UAV\footnote{Antigravity A1: \url{https://www.antigravity.tech}} with an integrated \(360^\circ\) camera.
We then sample candidate frames from these sequences and manually remove temporally adjacent frames with highly overlapping visual content and low-quality images. 
To balance the benchmark, we retain $250$ panoramic images per platform.

We annotate all selected images at the pixel level following the $19$-class Cityscapes semantic taxonomy~\cite{cordts2016cityscapes}.
A pretrained segmentation model is first used to generate an initial semantic mask for each image, which is subsequently inspected and corrected by human annotators following consistent annotation guidelines. Pixels belonging to categories outside the target taxonomy or lying in regions that cannot be reliably identified are assigned the ignore label. In total, EmbPASS contains $1,000$ annotated panoramic images.

\subsection{Dataset Statistics and Analysis}
The basic statistics of the platform-specific subsets are summarized in Fig.~\ref{fig:benchmark_characteristics}(a). 
\emph{Drone} and \emph{Wearable} cover the full spherical latitude range, whereas \emph{Vehicle} and \emph{Quadruped} cover most of the latitude range and a relatively narrow latitude band, respectively.
We further group the $19$ semantic classes into seven coarse semantic groups and compute the proportion of each group among all valid annotated pixels. 
As shown in Fig.~\ref{fig:benchmark_characteristics}(b), the relative proportions of different semantic groups vary across platforms.

Beyond semantic composition, we further analyze the spherical spatial organization of scene semantics. Specifically, we compute the normalized pixel distribution of each semantic class along the spherical latitude axis to characterize platform-dependent spherical semantic layouts. As shown in Fig.~\ref{fig:benchmark_characteristics}(c), although the coarse vertical semantic structure is broadly preserved, different semantic classes exhibit noticeable platform-dependent variations in their dominant locations and latitude ranges. For example, compared with \emph{Vehicle} and \emph{Wearable}, the distributions of ground-related and several movable object classes in \emph{Drone} shift toward more negative spherical latitudes, whereas those in \emph{Quadruped} are concentrated within a narrower near-equatorial observation range.
Since ERP projection distortion varies with spherical latitude, these cross-platform differences in semantic layout mean that scene content is observed under varying distortion conditions, increasing the geometric variability that cross-embodiment panoramic perception must accommodate.

\section{Proposed Method}
\begin{figure*}[!t]
    \centering
    \includegraphics[
        width=0.97\textwidth
    ]{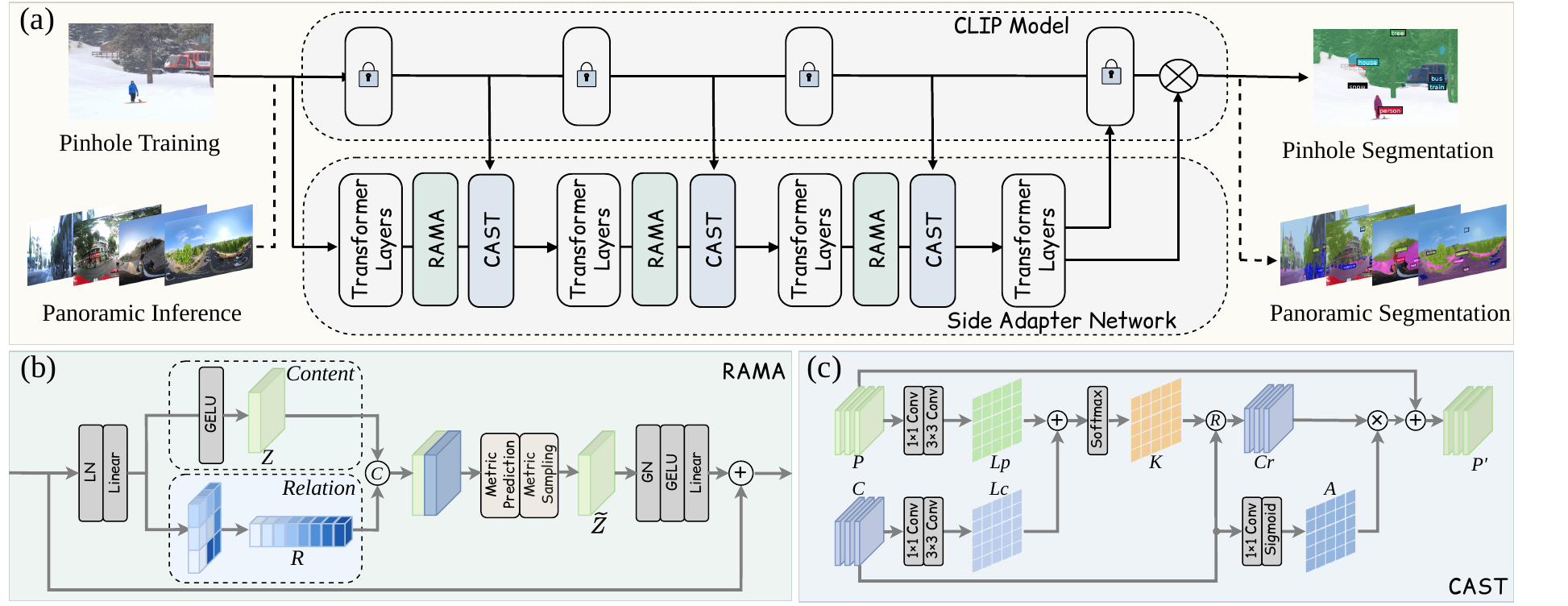}
    \vskip -1.5ex
    \caption{
        \hspace*{-0.5em}%
        \textbf{Overview of EPONet.}
        (a) Overall framework with a frozen CLIP model and a trainable Side branch.
        (b) Relation-Aware Metric Adapter (RAMA) for relation-aware metric prediction and structured spatial aggregation.
        (c) Content-Adaptive Semantic Transfer (CAST) for content-adaptive reassembly and residual transfer of pretrained CLIP features.
    }
    \label{fig:framework}
    \vskip -3ex
\end{figure*}

\subsection{Overall Framework}
\label{sec:overall}

We propose \textbf{EPONet}, an open-vocabulary panoramic semantic segmentation network for heterogeneous embodied platforms.
EPONet is trained only on pinhole images from a label-rich source domain and directly applied to panoramic observations captured by unseen embodied platforms.

As shown in Fig.~\ref{fig:framework}, the overall framework consists of a frozen CLIP visual encoder and a lightweight trainable Side Adapter Network~\cite{san}.  CLIP preserves the open-vocabulary recognition capability acquired from vision-language pre-training, whereas the Side branch comprises multiple transformer layers and RAMA modules for spatial modeling under heterogeneous panoramic observations. Meanwhile, CAST adaptively transfers intermediate CLIP representations to the Side branch to better exploit transferable pretrained semantics under varying panoramic observations. Sec.~\ref{sec:rama} and Sec.~\ref{sec:cast} detail the two modules, respectively. The resulting Side representations generate category-agnostic mask proposals and corresponding attention biases. Guided by these biases, the frozen CLIP encoder performs open-vocabulary recognition of the candidate regions. Finally, the mask proposals are combined with the category predictions to produce the pixelwise semantic segmentation map.

\subsection{Relation-Aware Metric Adapter}
\label{sec:rama}
Panoramic observations across heterogeneous embodied platforms expose scene content to varying distortion conditions, increasing the geometric variability that spatial modeling must accommodate. To address this challenge, we propose Relation-Aware Metric Adapter (RAMA), a lightweight residual adaptation module that structures local spatial aggregation through relation-aware local metrics.

\noindent\textbf{Metric-Guided Structured Sampling.}
Local deformations induced by panoramic projection vary across spatial positions and directions, while local metrics provide a structured modeling approach for such non-uniform geometric variations by jointly characterizing direction-dependent sampling scales and anisotropy. RAMA adopts the metric-guided structured sampling formulation of MetricConv~\cite{metricconv}, in which the local sampling geometry is parameterized using spatially varying metrics. Rather than predicting pointwise sampling offsets, this metric parameterization constrains the entire sampling support through a shared local geometry, thereby providing a structured geometric inductive bias for spatial modeling under varying panoramic distortion conditions.

At each spatial position \(p\), we parameterize the local metric \(F_p\) as a Randers metric to characterize both anisotropy and directional asymmetry of the local sampling geometry:
\begin{equation}
F_p(\mathbf{u}) =\sqrt{\mathbf{u}^{\top}M_p\mathbf{u}}+\boldsymbol{\omega}_p^{\top}\mathbf{u},
\end{equation}
where \(M_p\succ0\) controls the orientation, scale, and anisotropy of the local sampling geometry, while \(\boldsymbol{\omega}_p\) introduces directional asymmetry and satisfies \(\|\boldsymbol{\omega}_p\|_{M_p^{-1}}<1\) to ensure the positivity of the Randers metric.

We use a nine-point reference support consisting of the center and eight surrounding directions derived from a \(3\times3\) neighborhood. For each non-center reference point, let \(\mathbf{u}_k\) denote its Euclidean unit direction and \(\rho_k\) its radial coefficient. By the positive homogeneity of \(F_p\), the corresponding sampling vector is given by:
\begin{equation}
\mathbf{s}_{p,k} = \rho_k\frac{\mathbf{u}_k}{F_p(\mathbf{u}_k)}.
\end{equation}
The center sampling vector is fixed at zero. 

\noindent\textbf{Relation-Aware Metric Prediction.}
The above metric parameterization provides a shared geometric constraint over the entire sampling support, while the local metric still needs to be adaptively determined from the input. Panoramic projection induces direction-dependent local structural variations, and observation differences across heterogeneous embodied platforms further increase the need for local metric adaptation. Therefore, relying solely on the content representation may be insufficient to characterize neighborhood organization along different spatial directions. Unlike MetricConv~\cite{metricconv}, which predicts the local metric only from the content representation, 
RAMA further captures directional center--neighbor relations through feature similarities and combines the resulting relation features with the content representation to predict relation-aware local metrics.

Given the Side feature \(P\), RAMA first projects it into a compact feature space as follows:
\begin{equation}
Z_{\mathrm{pre}}=\phi_{\mathrm{in}}(P),\qquad
Z=\operatorname{GELU}(Z_{\mathrm{pre}}),
\end{equation}
where \(\phi_{\mathrm{in}}\) denotes a lightweight input projection. For each spatial position \(p\), let \(q_{p,k}\) denote its neighbor along the \(k\)-th predefined spatial direction, \(k=1,\ldots,8\). We compute the cosine similarity between the features at \(p\) and \(q_{p,k}\):
\begin{equation}
r_{p,k} =
\frac{
Z_{\mathrm{pre},p}^{\top}Z_{\mathrm{pre},q_{p,k}}
}{
\left\|Z_{\mathrm{pre},p}\right\|_2
\left\|Z_{\mathrm{pre},q_{p,k}}\right\|_2
}.
\end{equation}
The resulting similarities are arranged according to their spatial directions into \(R_p=[r_{p,1},r_{p,2},\ldots,r_{p,8}]\) and collected over all positions to form the relation features \(R\). We then combine the content representation \(Z\) with the relation features \(R\) to predict the local Randers metric parameters:
\begin{equation}
\left(M_p,\boldsymbol{\omega}_p\right) = \Pi\left( f_M([Z;R])_p,\, f_{\omega}([Z;R])_p \right),
\end{equation}
where \([\cdot;\cdot]\) denotes channelwise concatenation, \(f_M\) and \(f_{\omega}\) predict the raw metric and drift parameters, respectively, and \(\Pi\) maps them to valid Randers parameters. Specifically, \(\Pi\) parameterizes \(M_p\) to have strictly positive eigenvalues and normalizes the drift vector such that \(\|\boldsymbol{\omega}_p\|_{M_p^{-1}}<1\), ensuring a valid Randers metric.

The resulting sampling locations are converted into offsets relative to the canonical \(3 \times 3\) grid and used for deformable feature aggregation over \(Z\), producing the adapted representation \(\widetilde{Z}\). 
Finally, RAMA projects the aggregated feature back to the Side feature dimension and performs residual fusion:
\begin{equation}
P'=P+ \boldsymbol{\gamma}\odot \phi_{\mathrm{out}}(\widetilde{Z}),
\end{equation}
where \(\phi_{\mathrm{out}}\) denotes the output projection and \(\boldsymbol{\gamma}\) is a learnable channelwise residual scale.

\subsection{Content-Adaptive Semantic Transfer}
\label{sec:cast}
Heterogeneous embodied platforms introduce substantial cross-embodiment observation shifts, making it difficult for models to maintain reliable semantic representations on unseen platforms. To better exploit the general and transferable semantic knowledge retained by the frozen CLIP encoder, we propose Content-Adaptive Semantic Transfer (CAST). Rather than directly transferring intermediate CLIP features to the Side branch, CAST adaptively reorganizes and transfers the pretrained representations according to the current CLIP and Side features.

At each CAST stage, let $P$ denote the current Side feature and $C$ the corresponding intermediate CLIP feature. We first obtain compact representations $P_c=\phi_P(P)$ and $C_c=\phi_C(C)$ through lightweight \(1\times1\) projections. We then construct a content-dependent reassembly kernel from \(P_c\) and \(C_c\)~\cite{fade}.
Specifically, the two representations are encoded by a shared \(3\times3\) content encoder into kernel logits \(L_P=g(P_c)\) and \(L_C=g(C_c)\), which are then combined and normalized to obtain the reassembly kernel \(K\):
\begin{equation}
K=\operatorname{Softmax}\!\left(L_P+L_C\right), \qquad C_r=\mathcal{R}(C,K),
\end{equation}
where \(g(\cdot)\) denotes the shared content encoder and \(\mathcal{R}(\cdot)\) denotes dynamic reassembly of the CLIP feature \(C\).

In parallel, we estimate a spatial modulation weight from the intermediate CLIP feature to control the residual transfer:
\begin{equation}
A=\sigma\!\bigl(a(C)\bigr), \qquad P'=P+A\odot C_r,
\end{equation}
where \(a(\cdot)\) denotes a lightweight \(1\times1\) gate predictor. The reassembly kernel \(K\) determines how intermediate CLIP features are locally reorganized according to the current CLIP and Side representations, while \(A\) modulates their spatially varying contribution to the Side branch.

\section{Experiments}
\subsection{Datasets and Evaluation Metrics}
\noindent\textbf{Datasets.}
Following the OPS setting~\cite{ops}, we train the model on the pinhole image dataset COCOStuff-164k~\cite{cocostuff}.
COCOStuff-164k contains $164k$ images annotated with $171$ semantic categories, and we use all $118k$ images from its training split for model training. 
We evaluate cross-embodiment open panoramic segmentation on the proposed EmbPASS benchmark and further report results on two public panoramic segmentation benchmarks, including Stanford2D3D~\cite{stanford2d3d} for indoor scenes and DensePASS~\cite{densepass} for outdoor street scenes. 
Stanford2D3D contains $1,413$ evaluation images annotated with $13$ semantic categories, while DensePASS contains $100$ evaluation images annotated with $19$ semantic categories. 
For Stanford2D3D, we report average results over three cross-validation folds.

\noindent\textbf{Evaluation Metrics.}
We report mean Intersection-over-Union (mIoU) and open mean Intersection-over-Union (open mIoU)~\cite{openmiou} to comprehensively evaluate open-vocabulary segmentation performance. mIoU relies on strict matching between predicted and ground-truth categories, whereas open mIoU further considers the WordNet-based semantic similarity~\cite{wordnet} between predicted and ground-truth categories and assigns partial credit to semantically related predictions.

\begin{table*}[!t]
    \centering

    \caption{
        Comparison with open-vocabulary segmentation methods on EmbPASS.
        We report mIoU and open mIoU (\%).
        Avg. denotes the platform-balanced average over the four target platforms.
        \#Params. is the number of learnable parameters in millions. Best and second-best results are shown in bold and underlined, respectively.
    }
    \label{tab:main_coco_mp}
    \vskip-1ex

    \footnotesize
    \setlength{\tabcolsep}{4.2pt}
    \renewcommand{\arraystretch}{1.08}

    \resizebox{\textwidth}{!}{%
    \begin{tabular}{
        l|c|cc|cc|cc|cc|cc
    }
        \toprule

        \multirow{2}{*}[-0.6ex]{Method}
        & \multirow{2}{*}[-0.6ex]{\#Params.}
        & \multicolumn{2}{c|}{
            \platformicon{figs/car_icon.png}\hspace{0.10em}Vehicle}
        & \multicolumn{2}{c|}{
            \platformicon{figs/drone_icon.png}\hspace{0.10em}Drone}
        & \multicolumn{2}{c|}{
            \platformicon{figs/wearable_icon.png}\hspace{0.10em}Wearable}
        & \multicolumn{2}{c|}{
            \platformicon{figs/quad_icon.png}\hspace{0.10em}Quadruped}
        & \multicolumn{2}{c}{Avg.} \\

        \cmidrule(lr){3-4}
        \cmidrule(lr){5-6}
        \cmidrule(lr){7-8}
        \cmidrule(lr){9-10}
        \cmidrule(lr){11-12}

        & &
        mIoU & open mIoU
        & mIoU & open mIoU
        & mIoU & open mIoU
        & mIoU & open mIoU
        & mIoU & open mIoU \\

        \midrule

        CAT-Seg~\cite{catseg}
        & 59.44
        & 36.94 & 44.29
        & 21.80 & 31.00
        & \underline{38.35} & \underline{45.11}
        & 35.67 & 41.82
        & 33.19 & 40.56 \\

        S2-Corr~\cite{s2corr}
        & 24.15
        & 35.93 & 42.57
        & 22.29 & 31.02
        & 36.12 & 42.45
        & 32.08 & 38.83
        & 31.60 & 38.72 \\

        SAN~\cite{san}
        & 8.37
        & 39.03 & 45.05
        & 25.36 & 33.01
        & 35.68 & 42.70
        & 35.22 & 40.70
        & 33.82 & 40.37 \\

        OOOPS (w/o RERP)~\cite{ops}
        & 9.17
        & 39.23 & 45.30
        & \textbf{26.01} & \textbf{33.66}
        & 34.85 & 41.72
        & 35.56 & 40.90
        & 33.91 & 40.39 \\

        OOOPS (w/ RERP)~\cite{ops}
        & 9.17
        & \underline{40.08} & \underline{46.20}
        & 25.42 & \underline{33.26}
        & 36.23 & 42.73
        & \underline{37.14} & \underline{42.35}
        & \underline{34.72} & \underline{41.14} \\

        \midrule

        EPONet
        & 8.80
        & \textbf{40.78} & \textbf{46.93}
        & \underline{25.90} & \textbf{33.66}
        & \textbf{38.87} & \textbf{45.17}
        & \textbf{37.74} & \textbf{43.43}
        & \textbf{35.82} & \textbf{42.30} \\

        \bottomrule
    \end{tabular}%
    }

    \vspace{1em}

    \begin{minipage}[t]{0.48\textwidth}
        \vspace{0pt}
        \centering

        \captionof{table}{
            Comparison on existing panoramic benchmarks using
            mIoU and open mIoU (\%). 
        }
        \label{tab:public_panorama}

        \vspace{-2pt}

        \small
        \setlength{\tabcolsep}{4.5pt}
        \renewcommand{\arraystretch}{1.08}

        \resizebox{0.97\linewidth}{!}{%
        \begin{tabular}{@{}lcccc@{}}
            \toprule

            \multirow{2}{*}[-0.6ex]{Method}
            & \multicolumn{2}{c}{Stanford2D3D~\cite{stanford2d3d}}
            & \multicolumn{2}{c}{DensePASS~\cite{densepass}} \\

            \cmidrule(lr){2-3}
            \cmidrule(lr){4-5}

            & mIoU & open mIoU
            & mIoU & open mIoU \\

            \midrule

            CAT-Seg~\cite{catseg}
            & 42.33 & 48.43
            & \underline{48.32} & \underline{54.98} \\

            S2-Corr~\cite{s2corr}
            & \textbf{44.21} & \textbf{49.24}
            & 46.53 & 53.01 \\

            SAN~\cite{san}
            & 43.36 & 48.40
            & 47.77 & 53.89 \\

            OOOPS (w/o RERP)~\cite{ops}
            & 43.14 & 48.40
            & 48.30 & 54.35 \\

            OOOPS (w/ RERP)~\cite{ops}
            & 42.60 & 47.78
            & 47.27 & 53.26 \\

            \midrule

            EPONet
            & \underline{43.87} & \underline{48.96}
            & \textbf{49.15} & \textbf{55.22} \\

            \bottomrule
        \end{tabular}%
        }

    \end{minipage}
    \hfill
    \begin{minipage}[t]{0.48\textwidth}
        \vspace{0pt}
        \centering

        \captionof{table}{
            Ablation studies on EmbPASS:
            (a) model components and
            (b) CAST insertion stages without RAMA.
            Results are averaged across the four platforms.
        }
        \label{tab:component_cast_ablation}

        \vspace{-7pt}

        \small
        \setlength{\tabcolsep}{3pt}
        \renewcommand{\arraystretch}{1.08}

        \edef\componentrowgap{%
            \the\dimexpr0.36\baselineskip\relax
        }

        \begin{minipage}[t]{0.55\linewidth}
            \vspace{0pt}
            \centering

            (a) Components\par\vspace{2pt}

            \begin{tabular*}{\linewidth}{
                @{\extracolsep{\fill}}cccc@{}
            }
                \toprule
                RAMA & CAST & mIoU & $\Delta$ \\
                \midrule
            
                & & 33.82 & -- \\
                [\componentrowgap]
            
                $\checkmark$ & & 34.99 & +1.17 \\
                [\componentrowgap]
            
                & $\checkmark$ & 34.78 & +0.96 \\
                [\componentrowgap]
            
                $\checkmark$ & $\checkmark$
                & \textbf{35.82} & \textbf{+2.00} \\
            
                \bottomrule           
            \end{tabular*}
        \end{minipage}
        \hfill
        \begin{minipage}[t]{0.36\linewidth}
            \vspace{0pt}
            \centering

            (b) CAST stages\par\vspace{2pt}

            \begin{tabular*}{\linewidth}{
                @{\extracolsep{\fill}}cc@{}
            }
                \toprule
                Stages & mIoU \\
                \midrule

                $\{-\}$ & 33.82 \\
                $\{3\}$ & 34.64 \\
                $\{2,3\}$ & 34.68 \\
                $\{1,2,3\}$ & \textbf{34.78} \\
                $\{0,1,2,3\}$ & 34.48 \\

                \bottomrule
            \end{tabular*}
        \end{minipage}

    \end{minipage}

\vskip-4ex
\end{table*}

\subsection{Implementation Details}
We implement our model in PyTorch using Detectron2~\cite{detectron2} and use a frozen CLIP ViT-B/16~\cite{clip} as the visual encoder.
The model is trained on a single NVIDIA RTX 3090 GPU.
We use the AdamW optimizer with an initial learning rate of $1{\times}10^{-4}$ and a weight decay of $1{\times}10^{-4}$.
The batch size is $8$, and training lasts for $60k$ iterations using a polynomial learning-rate schedule with a power of $0.9$.
During training, the shorter side of each input image is randomly resized to $320{\sim}1024$ pixels and then randomly cropped to $640{\times}640$.
Mask recognition is supervised with cross-entropy loss, while mask proposal generation is supervised with binary cross-entropy loss and Dice loss.
During inference, we first resize the shorter side of each input image to $640$ pixels while limiting the longer side to at most $2560$ pixels.
We then perform sliding-window prediction with a window size of $640{\times}640$ and a stride of $320{\times}320$. 
For fair comparison, all baselines are retrained under the same training and inference settings described above.

\subsection{Comparison with State-of-the-Art Methods}
\noindent\textbf{Results on EmbPASS.}
We compare EPONet with representative open-vocabulary segmentation methods and the panorama-oriented OOOPS~\cite{ops}, with the results reported in Table~\ref{tab:main_coco_mp}. EPONet achieves the best platform-balanced performance, reaching $35.82\%$ mIoU and $42.30\%$ open mIoU. Compared with SAN~\cite{san}, EPONet improves the average mIoU and open mIoU by $2.00\%$ and $1.93\%$, respectively, with only $0.43M$ additional learnable parameters. Moreover, EPONet outperforms the strongest competing method OOOPS with RERP by $1.10\%$ mIoU and $1.16\%$ open mIoU while using $0.37M$ fewer learnable parameters.

Across the four acquisition platforms, our method achieves the best mIoU on \emph{Vehicle}, \emph{Wearable}, and \emph{Quadruped}, reaching $40.78\%$, $38.87\%$, and $37.74\%$, respectively. On \emph{Drone}, our method obtains $25.90\%$ mIoU, only $0.11\%$ below the best result, while tying for the best open mIoU of $33.66\%$.
Notably, all compared methods perform substantially worse on \emph{Drone} than on the other platforms, indicating that the aerial subset constitutes a particularly challenging panoramic observation setting. This trend is consistent with the elevated viewpoint and distinct spherical semantic layout of \emph{Drone}, where scene content is distributed toward higher absolute latitudes and consequently experiences stronger ERP projection distortion. Overall, our method achieves consistently strong performance across heterogeneous embodied platforms while obtaining the best platform-balanced results.

\begin{figure*}[t]
    \centering

    \newlength{\qualw}
    \newlength{\qualwquad}
    \newlength{\qualh}

    \setlength{\qualh}{0.0975\textwidth}

    \setlength{\qualw}{0.195\textwidth}

    \setlength{\qualwquad}{0.390\textwidth}

    \setlength{\tabcolsep}{0pt}
    \renewcommand{\arraystretch}{0}

    \begin{tabular}{
        @{}
        c@{\hspace{0.002\textwidth}}
        c@{\hspace{0.002\textwidth}}
        c@{\hspace{0.002\textwidth}}
        c
        @{}
    }

        \makebox[\qualw][c]{%
            {\scriptsize\sffamily\bfseries Vehicle}%
        }
        &
        \makebox[\qualw][c]{%
            {\scriptsize\sffamily\bfseries Drone}%
        }
        &
        \makebox[\qualw][c]{%
            {\scriptsize\sffamily\bfseries Wearable}%
        }
        &
        \makebox[\qualwquad][c]{%
            {\scriptsize\sffamily\bfseries Quadruped}%
        }
        \\[2pt]

        \includegraphics[
            width=\qualw,
            height=\qualh
        ]{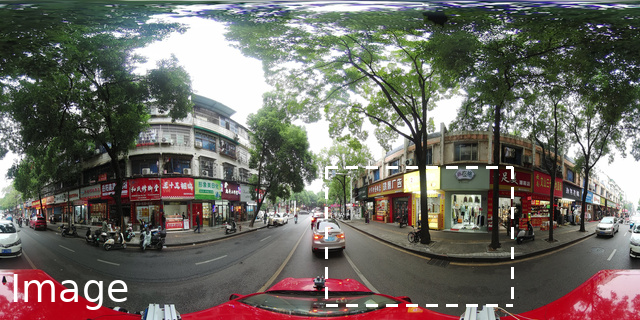}
        &
        \includegraphics[
            width=\qualw,
            height=\qualh
        ]{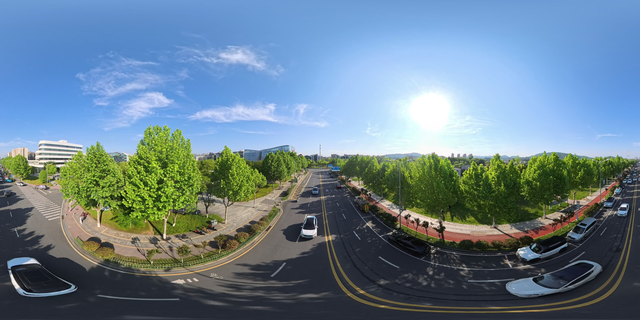}
        &
        \includegraphics[
            width=\qualw,
            height=\qualh
        ]{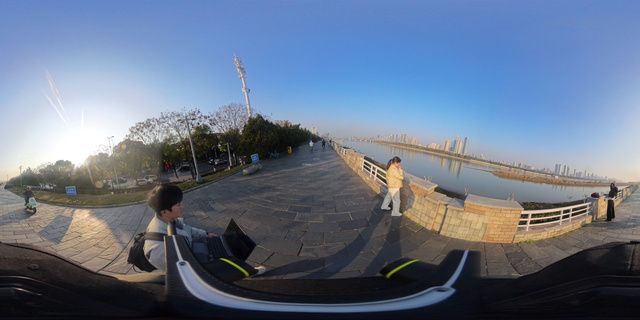}
        &
        \includegraphics[
            width=\qualwquad,
            height=\qualh
        ]{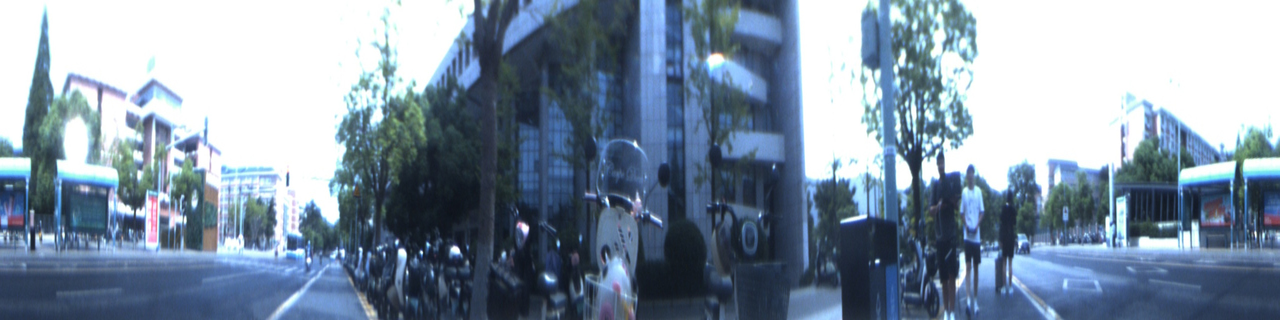}
        \\[0.8pt]

        \includegraphics[
            width=\qualw,
            height=\qualh
        ]{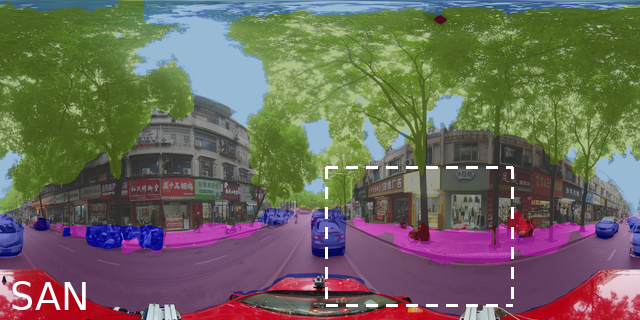}
        &
        \includegraphics[
            width=\qualw,
            height=\qualh
        ]{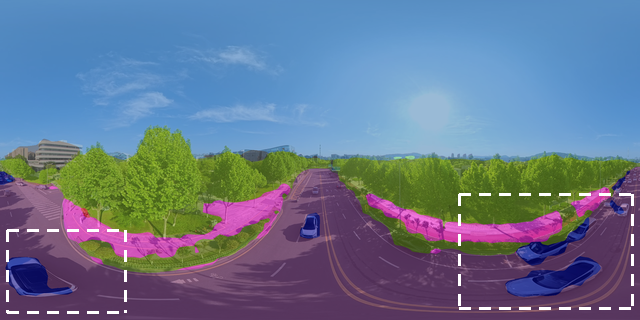}
        &
        \includegraphics[
            width=\qualw,
            height=\qualh
        ]{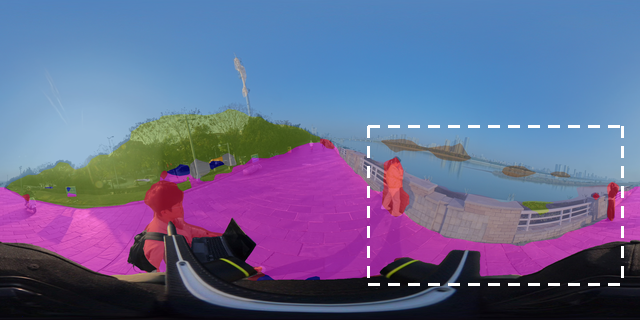}
        &
        \includegraphics[
            width=\qualwquad,
            height=\qualh
        ]{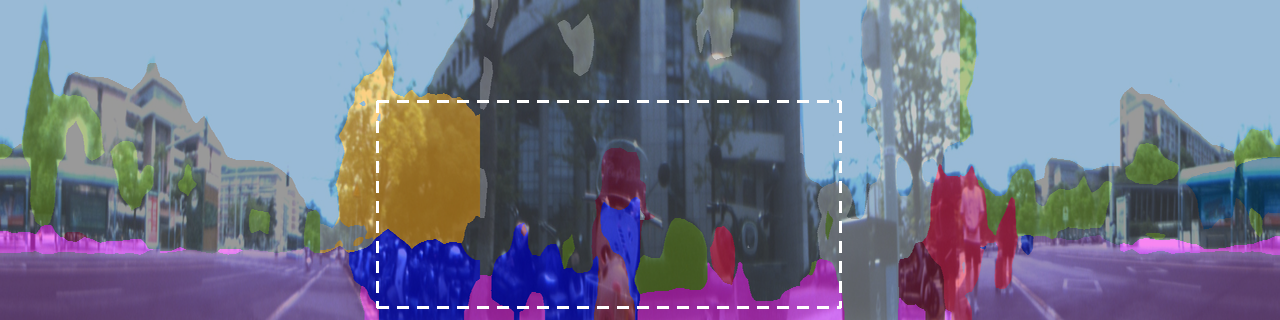}
        \\[0.8pt]

        \includegraphics[
            width=\qualw,
            height=\qualh
        ]{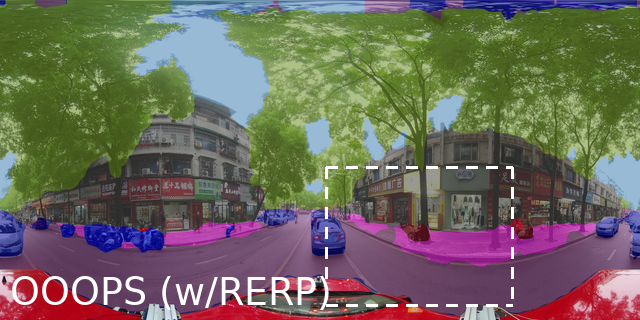}
        &
        \includegraphics[
            width=\qualw,
            height=\qualh
        ]{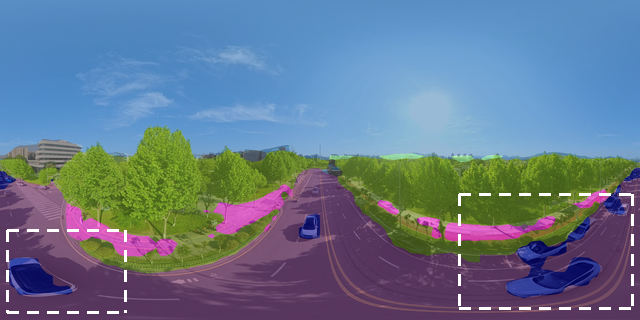}
        &
        \includegraphics[
            width=\qualw,
            height=\qualh
        ]{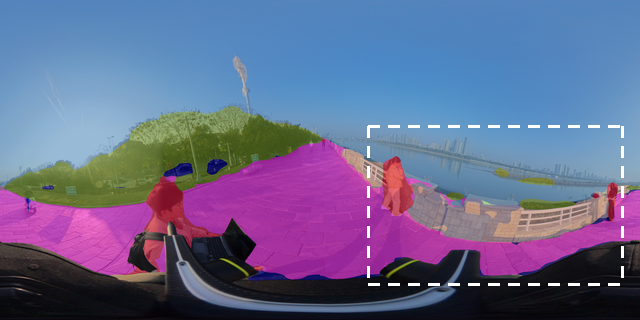}
        &
        \includegraphics[
            width=\qualwquad,
            height=\qualh
        ]{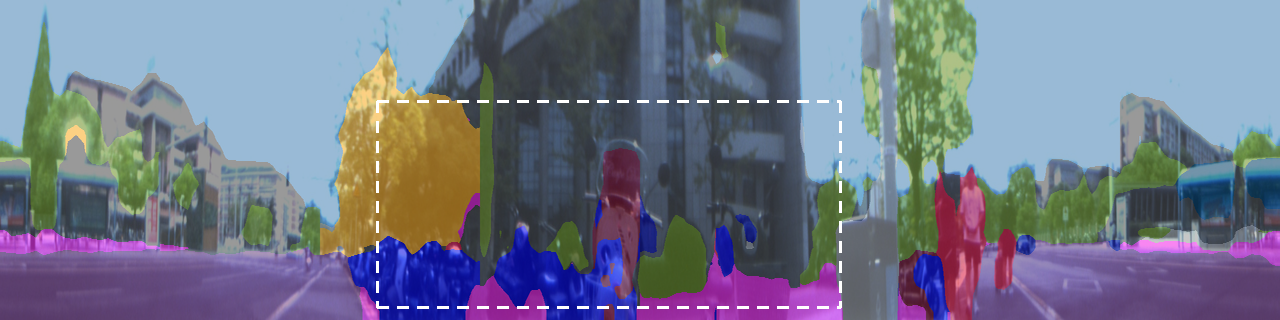}
        \\[0.8pt]

        \includegraphics[
            width=\qualw,
            height=\qualh
        ]{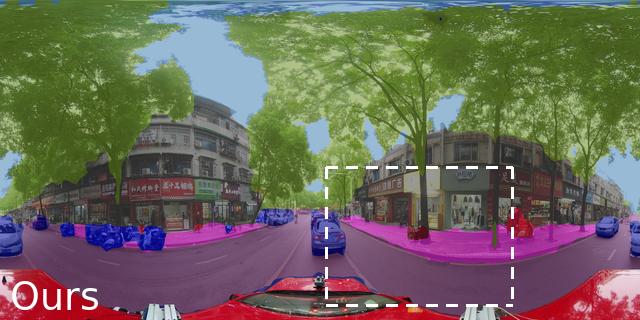}
        &
        \includegraphics[
            width=\qualw,
            height=\qualh
        ]{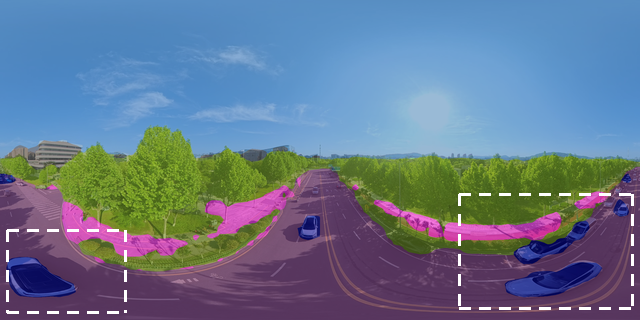}
        &
        \includegraphics[
            width=\qualw,
            height=\qualh
        ]{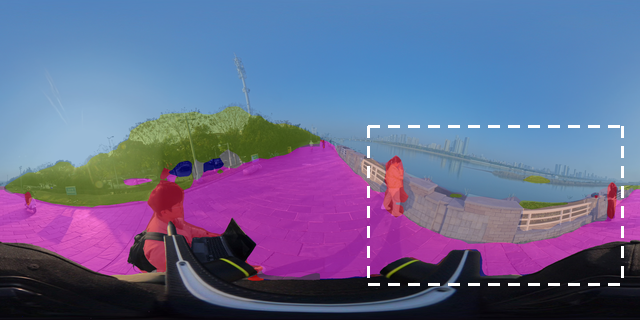}
        &
        \includegraphics[
            width=\qualwquad,
            height=\qualh
        ]{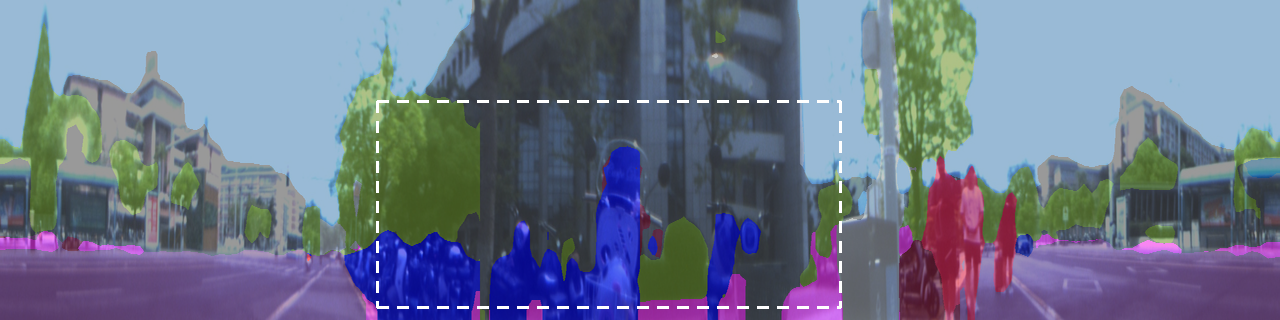}
        \\[0.8pt]

        \includegraphics[
            width=\qualw,
            height=\qualh
        ]{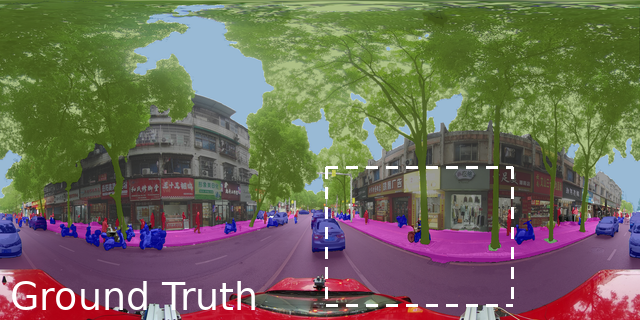}
        &
        \includegraphics[
            width=\qualw,
            height=\qualh
        ]{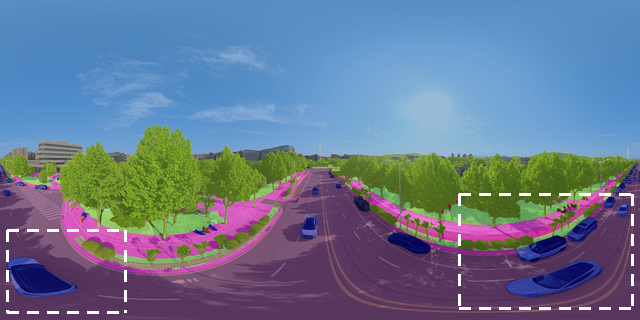}
        &
        \includegraphics[
            width=\qualw,
            height=\qualh
        ]{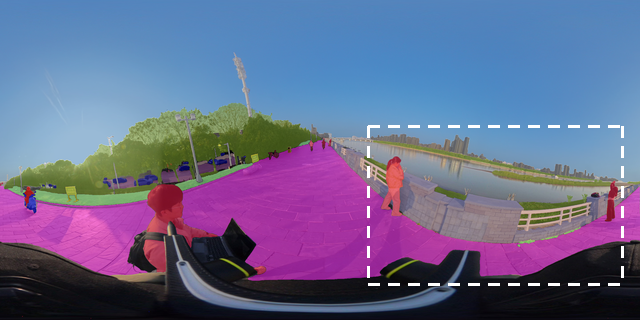}
        &
        \includegraphics[
            width=\qualwquad,
            height=\qualh
        ]{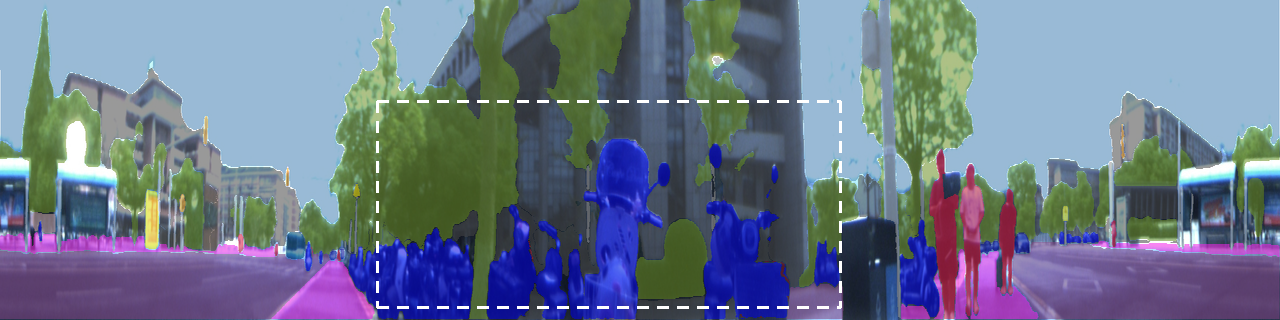}

    \end{tabular}
    \caption{
        \hspace*{-0.5em}%
        \textbf{Qualitative comparison on EmbPASS.}
        From top to bottom, we show the input image, predictions from SAN~\cite{san}, OOOPS with RERP~\cite{ops}, our method, and the ground truth. Dashed boxes highlight representative regions for visual comparison.
    }
    \label{fig:qualitative_comparison}
    \vskip-3ex
\end{figure*}

\noindent\textbf{Results on the Public Panoramic Benchmarks.}
Beyond the proposed EmbPASS benchmark, we further evaluate our method on Stanford2D3D~\cite{stanford2d3d} and DensePASS~\cite{densepass}, representing indoor and outdoor panoramic scenes, respectively. 
As shown in Table~\ref{tab:public_panorama}, our method achieves $43.87\%$ mIoU and $48.96\%$ open mIoU on Stanford2D3D, with the mIoU ranking second only to S2-Corr.
On DensePASS, our method achieves $49.15\%$ mIoU and $55.22\%$ open mIoU, obtaining the best performance on both metrics.

\subsection{Ablation Studies}

\noindent\textbf{Effects of Model Components.}
We first evaluate the individual contributions of RAMA and CAST. 
As shown in Table~\ref{tab:component_cast_ablation}(a), the base configuration achieves $33.82\%$ mIoU. Introducing RAMA or CAST individually improves the performance to $34.99\%$ and $34.78\%$, corresponding to gains of $1.17\%$ and $0.96\%$, respectively. These individual performance gains support the effectiveness of relation-aware metric adaptation and content-adaptive transfer of pretrained semantics for cross-embodiment panoramic perception. When both modules are employed, the mIoU further increases to $35.82\%$, improving over the base configuration by $2.00\%$ and indicating their complementary effects.

\begin{figure}[!t]
    \centering
    \includegraphics[
        width=0.98\columnwidth
    ]{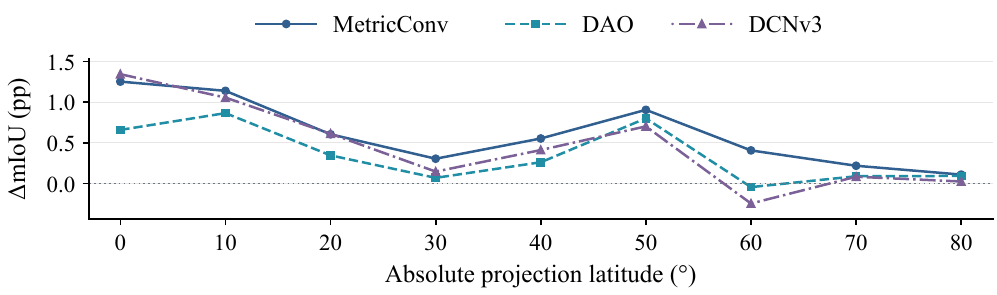}
    \vskip-2ex
    \caption{
        \hspace*{-0.5em}%
        \textbf{Controlled projection analysis across projection-center latitudes.}
        Positive $\Delta\mathrm{mIoU}$ favors RAMA.
    }
    \label{fig:controlled_projection}
    \vskip -3ex
\end{figure}

\noindent\textbf{Effect of Spatial Aggregation Operators.}
To further analyze the effect of different spatial aggregation strategies, we replace RAMA with several representative spatial operators while keeping the remaining network architecture unchanged. As shown in Table~\ref{tab:operator_ablation}, replacing RAMA with standard convolution yields $34.15\%$ mIoU, while DCN, DCNv2, and DCNv3 achieve $34.43\%$, $34.43\%$, and $34.46\%$, respectively. DAO and MetricConv achieve $34.09\%$ and $34.48\%$ mIoU, respectively. RAMA achieves the highest overall mIoU of $34.99\%$, outperforming MetricConv by $0.51\%$. This comparison highlights two aspects of our design. 
Compared with offset-based deformable operators, RAMA constrains the entire sampling support through shared local geometry rather than independently predicting pointwise sampling offsets. Compared with MetricConv, RAMA further incorporates directional center--neighbor relations into local metric prediction. The consistent improvements over both types of operators support the effectiveness of relation-aware metric adaptation for spatial modeling under heterogeneous panoramic observations.

\begin{table}[!t]
    \centering
    \caption{
        Comparison of spatial aggregation operators on EmbPASS.
        Results are averaged across the four platforms.
    }
    \vskip -1ex
    \label{tab:operator_ablation}
    \small
    \setlength{\tabcolsep}{6pt}
    \renewcommand{\arraystretch}{1.08}
    \begin{tabular}{lcc}
        \toprule
        Method & mIoU & Params. (M) \\
        \midrule
        SAN~\cite{san}
        & 33.82 & 8.37 \\
        \midrule
        $+\;$Conv
        & 34.15 & 8.65 \\
        $+\;$DCN~\cite{dcn}
        & 34.43 & 8.69 \\
        $+\;$DCNv2~\cite{dcnv2}
        & 34.43 & 8.71 \\
        $+\;$DCNv3~\cite{dcnv3}
        & 34.46 & 8.56 \\
        $+\;$DAO~\cite{ops}
        & 34.09 & 8.56 \\
        $+\;$MetricConv~\cite{metricconv}
        & 34.48 & 8.66 \\
        \midrule
        $+\;$RAMA
        & \textbf{34.99} & 8.66 \\
        \bottomrule
    \end{tabular}
    \vskip -3ex
\end{table}

We further evaluate segmentation under controlled panoramic projection conditions using Spherical Projection Transformation (SPT)~\cite{tan2025dascspt} on COCO validation images and annotations, with the projection-center latitude ranging from \(-80^\circ\) to \(80^\circ\) at \(10^\circ\) intervals. We compute mIoU over source regions jointly visible across all views and average the results over paired positive and negative latitudes. 
As shown in Fig.~\ref{fig:controlled_projection}, RAMA outperforms MetricConv~\cite{metricconv} at all evaluated latitudes and DAO and DCNv3 at most latitudes, with average gains of \(0.53\%\), \(0.31\%\), and \(0.35\%\), respectively, across all nonzero latitudes, further supporting the effectiveness of RAMA under diverse panoramic projection conditions.

\noindent\textbf{Effect of CAST Insertion Stages.}
We further investigate different stage configurations of CAST across feature interaction stages. 
As shown in Table~\ref{tab:component_cast_ablation}(b), applying CAST only at stage $\{3\}$ yields $34.64\%$ mIoU. Further applying CAST at stages $\{2,3\}$ and $\{1,2,3\}$ improves the performance to $34.68\%$ and $34.78\%$, respectively. 
These consistent improvements support the effectiveness of content-adaptive semantic transfer at intermediate feature interaction stages. When CAST is further extended to the earliest stage $\{0\}$, the performance decreases to $34.48\%$. 
This may be because shallow CLIP representations contain less high-level semantic information, making premature local reassembly less effective in exploiting pretrained semantics and potentially interfering with the spatial information required for panoramic modeling.
Therefore, we adopt CAST at stages $\{1,2,3\}$ in the final model.

\subsection{Qualitative Analysis}
As shown in Fig.~\ref{fig:qualitative_comparison}, we visualize the segmentation results of EPONet and representative baselines across the four embodied platforms. Compared with the baselines, EPONet produces more complete and spatially coherent segmentation results. As highlighted by the dashed boxes, EPONet better preserves the spatial continuity of semantic regions and the integrity of their boundaries, while producing cleaner predictions around foreground objects with fewer local misclassifications and fragmented regions. These qualitative results further indicate that EPONet yields more stable semantic predictions under heterogeneous observations, thereby supporting more reliable scene understanding for downstream robotic perception across diverse embodied platforms.

\section{Conclusion}
In this work, we introduce Cross-Embodiment Open Panoramic Segmentation, aiming to advance comprehensive panoramic scene understanding across heterogeneous embodied platforms. To this end, we establish EmbPASS, a multi-platform panoramic semantic segmentation benchmark spanning \emph{Vehicle}, \emph{Drone}, \emph{Wearable}, and \emph{Quadruped} platforms. We further propose EPONet, which integrates Relation-Aware Metric Adapter (RAMA) and Content-Adaptive Semantic Transfer (CAST) to enable effective cross-embodiment panoramic perception. Extensive experiments on EmbPASS and existing panoramic benchmarks demonstrate the effectiveness of EPONet, which achieves the best platform-balanced performance on EmbPASS while remaining competitive on Stanford2D3D and DensePASS. We hope this work will facilitate research toward unified and generalizable panoramic perception for embodied systems.

\bibliographystyle{IEEEtran}
\bibliography{bib}

\end{document}